\documentclass{article}
\pdfoutput=1 

\usepackage[preprint]{neurips_2026}

\usepackage[utf8]{inputenc}
\DeclareUnicodeCharacter{03C4}{\ensuremath{\tau}} 
\usepackage[T1]{fontenc}
\usepackage{hyperref}
\usepackage{url}
\usepackage{booktabs}
\usepackage{amsfonts}
\usepackage{nicefrac}
\usepackage{microtype}
\usepackage{xcolor}
\usepackage{wrapfig}
\usepackage{caption}

\usepackage{graphicx}
\usepackage{multirow}
\usepackage{amsmath}
\usepackage{array}
\usepackage{algorithm}
\usepackage{algpseudocode}
\usepackage{tikz}
\IfFileExists{inconsolata.sty}{\usepackage{inconsolata}}{} 
\usepackage{latexsym}

\title{Agentic Systems Need Multi-Run Evaluation:
A Study of Retrieval and Response Variability in a Deployed Enterprise RAG System}

\author{
  Karan Gupta \\
  Adobe \\
  Noida, India \\
  \texttt{karagupta@adobe.com}
  \And
  Sumit Bhatia \\
  Adobe \\
  Noida, India \\
  \texttt{sumbhati@adobe.com}
  \And
  Yunyao Li \\
  Adobe \\
  San Jose, USA \\
  \texttt{yunyaol@adobe.com}
}

\begin{document}

\maketitle

\begin{abstract}
Previous RAG systems have mostly been evaluated in a single-run setting, and with the
advent of agentic RAG, systems are still often evaluated the same way. However, agentic RAG systems introduce a source of run-to-run variability, since several of their components make decisions non-deterministically. 
Such systems typically comprise a retrieval stage followed by a generation stage, and the retrieval stage in particular can behave differently across runs. For example, the agent might issue
a different number of parallel searches from one run to the next, which alters the retrieved documents and therefore the final answer.
In this paper we study this behavior in a deployed agentic RAG system for
product-documentation questions, running each of
739 questions ten times and analyzing the execution traces to find where different runs
first diverge. We find a substantial gap between success in any run and success in every
run: the retrieval hit rate is 69.4\% when the correct page is found in at least one run,
but 50.9\% when it must be found in all ten. Responses also stay similar in meaning while
varying in wording, with an average BERTScore of 0.90 and a BLEU of 0.49. Trace-level
analysis shows that retrieval is the largest source of divergence (43.9\%), and the number
of parallel searches at the first retrieval step alone accounts for 20.1\% of divergent
pairs. Taken together, these results show that single-run evaluation can hide reliability
and consistency issues in agentic RAG systems, and that multi-run evaluation is necessary to reveal them.
\end{abstract}

\section{Introduction \& Related Work}
Retrieval-augmented generation (RAG) is the standard way to provide large language
models with grounding information when their pre-trained knowledge is not sufficient
\citep{3495724.3496517}. Agentic RAG systems add autonomous steps to the retrieval
layer, where the model itself decides what to search for, how many searches to issue,
and when it has gathered enough information to stop \citep{ac1f09077393404a8bea5141d8710259,
singh2026agenticretrievalaugmentedgenerationsurvey}. Because these decisions are sampled
from the model \citep{DBLP:journals/corr/abs-1904-09751}, they can change from one run to
the next, which introduces a natural variability into the results. The same query,
executed twice, can follow different retrieval steps and arrive at a different answer.

Agentic RAG systems are currently evaluated in the same way as static RAG systems, with
a single execution per query and a single reported score. RAG evaluation tools such as
RAGAS \citep{es-etal-2024-ragas}, ARES \citep{saad-falcon-etal-2024-ares}, and
RAGChecker \citep{3737916.3738608} score dimensions such as faithfulness, relevance, and
groundedness, and agent benchmarks such as AgentBench
\citep{DBLP:journals/corr/abs-2308-03688}, WebArena
\citep{zhou2024webarenarealisticwebenvironment}, and SWE-bench
\citep{ae6ae0c3765a4f9dbba578819d5b59b0} measure task completion. Neither, however,
measures whether a system's output is stable across repeated runs. More broadly, the
limitations of evaluating a model using a single input have also been observed in LLM
evaluation. \citet{mizrahi-etal-2024-state} show that semantically equivalent instruction
templates can lead to substantially different performance and even different model
rankings, and argue for evaluating models across multiple instruction templates.
Similarly, \citet{chatterjee-etal-2024-posix} propose POSIX, a prompt-sensitivity index
that measures how model behavior changes under intent-preserving prompt variations.
These studies vary the input while keeping the underlying task fixed. Agentic systems
introduce a complementary source of variability: even when the input prompt is fixed,
the system can make different intermediate decisions across executions. A single run
therefore does not indicate whether a reported success would recur, and the variance
across runs can be large enough that an apparent improvement falls within the noise
\citep{rabanser2026scienceaiagentreliability, Madaan2024QuantifyingVI}. Evaluating an
agentic system reliably therefore requires looking across multiple runs rather than a
single one. Broader surveys of RAG \citep{gao2023ragsurvey} and agent evaluation
\citep{yehudai2025survey} provide a wider view of these evaluation settings.

Prior work has begun to study this run-to-run behavior. When an agentic system is
executed repeatedly and required to succeed consistently, as measured by pass$^{k}$
\citep{26dddae50d5a4610bdaced3e9fdb1c89}, its measured performance drops sharply relative
to a single run. \citet{mehta2026agentsdisagreethemselvesbehavioral} further show that
runs of such systems tend to diverge at their early steps. These studies establish that
inconsistency exists and roughly when it begins, but they do not identify the exact step
at which two runs diverge, nor how much of the overall variance each step of the agentic
pipeline accounts for. Answering this requires attributing the divergence between runs to
specific steps in the execution.

In this paper we study these questions on a deployed agentic RAG system. We execute each
query multiple times, align the resulting execution traces, and attribute the divergence
between runs to the step at which it first arises. We make the following contributions.
First, we separate \emph{any-run} accuracy, whether a query succeeds on at least one run,
from \emph{every-run} accuracy, whether it succeeds on all runs, and show that the two
differ substantially. Second, we measure cross-run consistency and find that responses
remain similar in meaning while differing in wording. Third, we introduce a trace-level
attribution that locates the step at which runs first diverge and quantifies the share of
divergence contributed by each step, and we find that divergence concentrates in the
retrieval-planning steps.

Consistency across runs has mostly been treated as a means rather than an object of study. Self-consistency \citep{DBLP:conf/iclr/0002WSLCNCZ23}, for example, samples multiple reasoning paths to select a final answer, rather than measuring the variability across those paths. Unlike single-run trace analyses such as ReAct \citep{ac1f09077393404a8bea5141d8710259} and Reflexion \citep{3666122.3666499}, we align traces across runs to identify where and how they differ. Adaptive-retrieval approaches such as Self-RAG \citep{ICLR2024_25f7be96} and active retrieval \citep{jiang-etal-2023-active} focus on learning when to retrieve and when to stop, whereas we study the run-to-run variability that remains when these decisions are sampled from the model. Our analysis requires only repeated runs and the execution traces the system already produces.

\section{Experimental Setup}
\label{sec:framework}

We study a deployed agentic RAG system that answers product-documentation questions. For a given
query, the system retrieves documents and then generates an answer, and it controls
its own retrieval, including what to search for, how many queries to issue, and how many
to run in parallel. We record the ordered sequence of steps taken in a single execution
as its \emph{trace}. Because the system's decisions are sampled, the same query can
produce different traces, and different answers, across runs.

A single execution is therefore not representative of the system, so we evaluate it over
repeated runs. We use $739$ questions, each paired with a ground-truth documentation page
(a GT URL) that answers it, and we run each question ten times, for $7{,}390$ runs in
total. The model, tools, document collection, and decoding settings are held fixed across the ten runs,
so we attribute the observed variation primarily to sampling. We take the ten runs of a question as our unit
of analysis. Because the approach needs only repeated runs and the traces the system
already logs, it is designed to be applicable across agentic RAG systems. We also evaluate a non-agentic
baseline, which retrieves once and generates an answer, to measure what the agentic
components add. For blind review, product names and other identifying information are omitted. Representative examples
of the questions and run-to-run divergences are provided in
Appendix~\ref{app:examples}

We measure retrieval reliability, answer quality, and cross-run consistency. A run counts
as a \emph{Hit} when the GT URL is among the pages it retrieves \citep{manning2008introduction}.
\emph{Per-run} accuracy is the Hit rate over all $7{,}390$ runs. \emph{Any-run} accuracy
treats a question as solved when the GT URL is retrieved on at least one of its ten runs,
and \emph{every-run} accuracy only when it is retrieved on all ten
\citep{Madaan2024QuantifyingVI, 26dddae50d5a4610bdaced3e9fdb1c89}. Any-run and every-run
give the best- and worst-case rates, and the gap between them reflects how much retrieval
varies from run to run. As Hit checks only retrieval, we also judge each answer with an
LLM \citep{3666122.3668142}. We split the answer into atomic claims
\citep{min-etal-2023-factscore} and rate each claim for \emph{correctness}, whether it is
factually accurate, and \emph{groundedness}, whether the retrieved documents support it
\citep{es-etal-2024-ragas, 3737916.3738608}. We rate \emph{comprehensiveness}, how
completely the response answers the question, over the full response rather than per
claim. For consistency, we take the
$\binom{10}{2}=45$ pairs of responses to a question and score each pair with BERTScore
\citep{DBLP:journals/corr/abs-1904-09675} for meaning and BLEU
\citep{papineni-etal-2002-bleu} for wording, reporting the mean, minimum, and maximum
\citep{mehta2026agentsdisagreethemselvesbehavioral}. When BERTScore is high but BLEU much
lower, the system expresses the same meaning in different words, which we call the
\emph{behavioral consistency gap}. The results of answer-quality comparison and cross-run consistency analysis are presented in Section~\ref{sec:setup}.

Finally, we attribute where two runs of a query diverge. We view an execution as a
sequence of decision steps, so the procedure is not tied to our particular system. For
each of the $45$ response pairs, an LLM judge first checks whether the two answers are
equivalent, meaning that they convey the same answer \citep{openai2024gpt4ocard,
bulian2022tomayto}, and we set the equivalent pairs aside. For a pair that differs, we align the two traces and identify the first step at which
the runs act differently. We then attribute the pair to that source of divergence:
\emph{routing} when the selected tools or document collections differ,
\emph{retrieval} when the search behavior differs, \emph{reasoning} when the
intermediate reasoning differs, \emph{trace depth} when one run performs additional
retrieval steps, and \emph{generation} when the retrieval pipeline is identical but
the final text differs. We compare the search queries and reasoning blocks of the two runs using TF cosine
similarity \citep{salton_termweighting_1988, manning2008introduction}. When multiple
queries are issued at a retrieval step, we align the queries across runs and use the
least similar matched pair to identify divergence. The detailed matching procedure and
thresholds are provided in Appendix~\ref{app:divergence}.

\section{Results}
\label{sec:setup}

\begin{wraptable}{r}{0.5\linewidth}
\centering\small
\setlength{\tabcolsep}{5pt}
\vspace{-\intextsep}
\begin{tabular}{lrr}
\toprule
Divergence type & Pairs & \% \\
\midrule
Routing & 476 & 5.9 \\
\textit{Retrieval (all rounds)} & \textit{3{,}555} & \textit{\textbf{43.9}} \\
\quad round-1 parallel count & 1{,}623 & 20.1 \\
Reasoning (all steps) & 1{,}374 & 17.0 \\
Trace depth & 990 & 12.2 \\
Generation & 1{,}454 & 18.0 \\
Miscellaneous & 243 & 3.0 \\
\midrule
Total divergent pairs & 8{,}092 & 100.0 \\
\bottomrule
\end{tabular}
\caption{Distribution of the $8{,}092$ divergent response pairs across pipeline steps,
each assigned to the first step at which the two runs act differently. Definitions in
Appendix~\ref{app:divergence}.}
\label{tab:divergence}
\end{wraptable}
\textbf{Retrieval reliability:} Retrieval accuracy depends strongly on how it is aggregated across runs. The retrieval
hit rate is $69.4\%$ under any-run accuracy and $50.9\%$ under every-run accuracy, a gap
of $18.5$ percentage points, with a per-run rate of $61.5\%$ in between. This gap
reflects considerable run-to-run variability in retrieval, and no single number is a
reliable summary on its own. A deployment that depends on repeated correct retrieval
cannot be judged from one execution.

\textbf{Cross-run consistency:} Cross-run consistency shows a similar pattern. Averaged over the $45$ response pairs per
question, mean BERTScore is $0.90$ and mean BLEU is $0.49$. Across questions, the per-question average BERTScore
ranges from $0.86$ to $0.94$, while BLEU ranges from $0.34$ to $0.69$. The high BERTScore shows
that repeated runs typically convey the same answer, while the moderate BLEU shows that
the exact wording changes from one run to the next. We refer to this dissociation as the
\emph{behavioral consistency gap}. In practice, it means that surface-level checks such
as exact-match or $n$-gram comparisons between runs are not reliable signals of a change
in behavior.

We group divergence into five main categories: \emph{routing}, when runs select
different tools or document collections; \emph{retrieval}, when they differ in the
number, type, or content of searches; \emph{reasoning}, when their reasoning differs
before or after retrieval; \emph{trace depth}, when one run takes additional steps or
retrieval rounds; and \emph{generation}, when the retrieval pipeline is otherwise the
same but the final response differs. Pairs that do not fall into these categories are
classified as \emph{miscellaneous}. The detailed detection rules and thresholds for these categories, and the full breakdown of the divergence
categories and their sub-categories are reported in  Appendix~\ref{app:divergence}.

Table~\ref{tab:divergence} attributes each divergent pair of runs to the step at which
the two runs first act differently. Retrieval is the largest source and explains $43.9\%$
of divergent pairs, and its dominant sub-cause is the number of parallel queries issued
at the first retrieval step, which alone explains $20.1\%$. Reasoning, trace depth, and
routing account for most of the remainder. Generation is the residual, in which the
retrieval pipeline is identical and only the final text differs, and accounts for
$18.0\%$; since it cannot be tied to an earlier step, we exclude it when identifying
where in the pipeline runs diverge. Among divergences that can be tied to a specific
step, retrieval accounts for $53.5\%$. Retrieval planning is thus the main cause of
inconsistency across runs.

For example, for the query \emph{please describe the type of admin and usage reporting
in Product~B}, one run issued four parallel searches at the first retrieval round,
including searches about audit logs, reporting analytics, 
while another issued three searches focused on admin reporting and report types. Thus,
the same question led to different retrieval plans at the first retrieval round.

\begin{wraptable}{r}{0.6\linewidth}
\centering\small
\setlength{\tabcolsep}{6pt}
\vspace{-\intextsep}
\begin{tabular}{lccc}
\toprule
Dimension & Non-agentic & Agentic & $\Delta$ \\
\midrule
Retrieval hit & 61.8 & 61.5 & $-0.3$ \\
Correctness & 82.8 & 80.9 & $-1.9$ \\
Groundedness & \textbf{95.7} & 82.9 & $-12.8$ \\
Comprehensiveness & 74.4 & \textbf{97.0} & $+22.6$ \\
\bottomrule
\end{tabular}
\caption{Retrieval hit rate and LLM-judged answer quality (\%); $\Delta$ is agentic minus
baseline. Retrieval hit rates are nearly equal, so the quality differences reflect answer
generation rather than retrieval.}

\label{tab:agentic}
\end{wraptable}
The agentic and non-agentic systems retrieve the ground-truth page at nearly the same
per-run rate ($61.5\%$ and $61.8\%$), so the differences in answer quality are not
explained by differences in retrieval hit rate alone (Table~\ref{tab:agentic}). Among runs that retrieve the GT page, $89.8\%$ continue searching after the hit
(Appendix~\ref{app:search}), indicating that the agent often performs further retrieval
after finding the relevant page. The agentic system is much
more comprehensive than the baseline ($+22.6$ points), so its answers cover more of what
each question asks. Its claims, however, are less grounded in the cited documents
($-12.8$) and marginally less accurate ($-1.9$). The agentic system therefore trades
grounding for coverage. A single aggregate quality score would hide this trade-off.

\section{Discussion}
\label{sec:practice}

Our central finding is that an agentic RAG system cannot be characterized by a single run or a single aggregate score, and that reliable evaluation instead requires repeated execution. Such systems therefore demand a different evaluation stance than the static pipelines they extend, for three reasons.

\begin{enumerate}
\item \textbf{Reliability is distinct from single-run accuracy.} A per-run score lies
somewhere between the any-run and every-run rates without indicating where, so systems
with the same reported score can differ substantially in run-to-run reliability.

\item \textbf{Consistency is multi-dimensional.} A system can remain stable in meaning while varying considerably in surface form, and different similarity metrics can therefore lead to different conclusions.

\item \textbf{Divergence originates in retrieval planning rather than generation.}
Although generation is often considered the primary source of stochasticity in language model systems, our analysis shows that divergence more often originates from earlier agent decisions. In particular, the number of parallel queries issued during the first retrieval step is the most common source of divergence, while generation accounts for a smaller fraction. This suggests that part of the observed variability is introduced by the agent's planning process rather than being an unavoidable property of final response generation.Planning decisions that lead to divergence could potentially be constrained through
system design, for example by making them deterministic functions of the query, but
the effect of such interventions remains to be evaluated. Generation-level variability is more difficult to eliminate and remains after earlier stages are stabilized.
\end{enumerate}

The comparison with the non-agentic baseline shows that improvements in comprehensiveness can occur together with lower grounding. These results suggest that aggregate scores can hide important differences between systems and may lead to different conclusions about which system performs better. The generation-level divergences include 185 cases where two runs produce directly contradictory factual claims despite following the same pipeline(see Appendix~\ref{app:divergence}). Unlike differences in wording or paraphrasing, these cases can provide conflicting information to the user without any difference in the retrieval trace. This shows that run-to-run variability is not simply evaluation noise. Repeated execution measures the extent of this variability, while trace-level analysis helps identify whether it originates from decisions that can be addressed through system design or from generation itself.

\section{Conclusion}
 
We presented a multi-run evaluation of a deployed agentic RAG system. Building on prior
work showing that repeated runs of an agent diverge, we attribute each divergence to
the specific pipeline step at which it first arises. Applied to $739$ questions, each
run ten times, the analysis makes the system's run-to-run behavior measurable. It
identifies retrieval planning as the step where most divergence originates, and it
shows that different quality dimensions can shift in opposite directions across
architectures.
 
The broader point is that a stochastic agentic system cannot be characterized by a
single execution, or by a single aggregate metric. Two directions follow for future
work. One is to test whether constraining the retrieval-planning decisions we identify
reduces inconsistency without harming answer quality. The other is to reproduce this
analysis on public benchmarks and across other agentic architectures. Limitations are
discussed in section~\ref{app:limitations}.

\section{Limitations}
\label{app:limitations}

Our reliability metric uses the GT URL as a proxy for a correct answer. A response can
be correct while relying on a different documentation page, and an incorrect response
can still retrieve the correct page, so Hit can miss real successes and count false
ones. The LLM judge scores each response directly, reducing this dependence. Second,
we evaluate a single deployed system in a single documentation question-answering
setting, so the exact percentages may not transfer to other systems even if the failure
modes we identify do. Third, we run each query ten times, which suffices to estimate
any-run and every-run rates but limits precision for rare events
\citep{miller2024addingerrorbarsevals}. Fourth, the comparison between the agentic and
non-agentic systems uses a GPT-4o judge rather than human labels. We use the same
judging procedure for both systems but do not independently validate the judge
against human annotations. The reported differences in comprehensiveness and
groundedness should therefore be interpreted with this limitation in mind. Fifth, we
attribute run-to-run variation to language-model sampling, but some non-determinism
can also arise from inference infrastructure \citep{he2025nondeterminism}. Finally, our
divergence attribution uses fixed TF-cosine similarity thresholds to keep classification
consistent across run pairs. We do not evaluate how attribution changes under
alternative thresholds, so the proportions of individual divergence categories may
depend on these settings. The taxonomy should therefore be viewed as a reproducible
attribution framework, with category proportions specific to the thresholds used
in our analysis.

\bibliographystyle{acl_natbib}
\bibliography{custom}

\appendix

\section{Divergence Attribution Details}
\label{app:divergence}

The divergence attribution procedure follows the trace-based analysis described in
Section~\ref{sec:framework}. For each non-equivalent pair, we align the execution
traces and identify the first step at which the two runs act differently.

\textbf{Trace structure:} We view an execution as a sequence of decision steps.
A divergence is classified as \emph{routing} when the selected tools or document
collections differ, \emph{retrieval} when the search behavior differs,
\emph{reasoning} when the intermediate reasoning differs, and \emph{trace depth}
when one run performs additional retrieval steps. When the retrieval pipeline is
identical but the final text differs, we classify the pair as \emph{generation}.
These categories provide the basis for the divergence taxonomy in
Table~\ref{tab:taxonomy}.

\textbf{Similarity computation:} We compare the search queries and reasoning
blocks of the two runs using TF cosine similarity
\citep{salton_termweighting_1988, manning2008introduction}. Each search query is
represented using a TF vectorizer with an $n$-gram range of $[1,1]$, i.e., unigrams
only, without IDF weighting. Reasoning blocks are compared in the same way over
their text.

\textbf{Matching procedure:} When two runs issue the same number of parallel
queries at a retrieval step, we align the queries using greedy one-to-one matching.
The most similar pair is matched first, after which those queries are removed and
the process is repeated. We use the similarity of the least similar matched pair as
the divergence signal. The procedure is summarized in
Algorithm~\ref{alg:match}.


\begin{wrapfloat}{algorithm}{r}{0.48\columnwidth}
\centering
\caption{Worst-matched query similarity}
\label{alg:match}
\begin{algorithmic}[1]
\Require query sets $A=\{a_1,\dots,a_m\}$, $B=\{b_1,\dots,b_m\}$
\State $S_{ij} \gets \cos(\mathrm{tf}(a_i), \mathrm{tf}(b_j))\ \forall i,j$
\State $M \gets \emptyset$
\While{$A$ and $B$ are both non-empty}
  \State $(i,j) \gets \arg\max_{i,j} S_{ij}$
  \State $M \gets M \cup \{(i,j)\}$; remove $a_i$ from $A$, $b_j$ from $B$
\EndWhile
\State \Return $\min_{(i,j)\in M} S_{ij}$
\end{algorithmic}
\end{wrapfloat}

\textbf{Thresholds:} We use fixed similarity thresholds to identify content and
reasoning divergence consistently across run pairs. Content divergence is flagged
when the worst-matched query similarity is below $0.40$, while reasoning divergence
is flagged when the reasoning-block similarity is below $0.45$. Pairs that pass all
per-step tests but have full-trace query similarity in $[0.40,0.85]$ are classified
as subtle wording divergence. Pairs with similarity above $0.85$ and no other
detected difference are attributed to generation. These thresholds provide a
consistent operationalization of the divergence categories. They are not intended
as universally calibrated cutoffs, so the reported proportions should be interpreted
in the context of this operationalization.

\textbf{Divergence taxonomy:} Table~\ref{tab:taxonomy} gives the detailed
taxonomy used to assign divergent pairs to the categories described above, including
the detection method and operational definition for each sub-category.

\begin{table*}[!htbp]
\centering\small
\begin{tabular}{p{3.6cm}p{4.4cm}p{5.0cm}}
\toprule
\textbf{Type} & \textbf{Detection method} & \textbf{Definition} \\
\midrule
Routing & Set equality on loaded configuration names & Runs load different tools or document collections. \\
Retrieval-1 parallel count & Count of parallel queries at round 1 & Different number of parallel queries in the first retrieval round. \\
Retrieval-1 query content & Greedy bipartite TF match, worst pair $<0.40$ & Same count, substantially different query content. \\
Pre-retrieval reasoning & TF cosine $<0.45$ on pre-retrieval reasoning & Same routing; reasoning diverges before any retrieval. \\
Post-retrieval-1 reasoning & TF cosine $<0.45$ on round-1 reasoning & Same round-1 retrieval; reasoning over the results diverges. \\
Retrieval-2 parallel count & Count of parallel queries at round 2 & Different number of parallel queries in the second retrieval round. \\
Retrieval-2 tool type & Tool-name comparison at round 2 & Same count, different tool (e.g.\ index vs.\ web fetch). \\
Retrieval-2 query content & Greedy bipartite TF match, worst pair $<0.40$ & Same count and tool, different query content. \\
Post-retrieval-2 reasoning & TF cosine $<0.45$ on round-2 reasoning & Reasoning over round-2 results diverges. \\
Trace depth & Total step-count comparison & One run performs additional retrieval rounds before answering. \\
Subtle wording (reclassified) & Full-trace query similarity in $[0.40,0.85]$ & Cumulative phrasing differences with no single clearly divergent step. \\
Miscellaneous & Later rounds and steps; subagent & Round 3+ retrieval, step 5+ reasoning, or subagent differences. \\
Generation & No retrieval on either side, or full-trace similarity $\geq 0.85$ & Identical pipeline; difference arising from generation. \\
\bottomrule
\end{tabular}
\caption{Full divergence taxonomy with the detection method for each type.}
\label{tab:taxonomy}
\end{table*}

The full distribution of divergent pairs is reported in
Table~\ref{tab:divergence_full}. The table expands the higher-level categories
reported in the main paper into their finer-grained retrieval and reasoning
sub-categories. In total, there are $8{,}092$ divergent pairs, of which
$3{,}555$ are attributed to retrieval, $1{,}374$ to reasoning, $990$ to trace
depth, and $1{,}454$ to generation.

\begin{table}[!htbp]
\centering\small
\begin{tabular}{lrr}
\toprule
Divergence type & Pairs & \% \\
\midrule
Routing & 476 & 5.9 \\
\addlinespace
\textit{Retrieval (all rounds)} & \textit{3{,}555} & \textit{43.9} \\
\quad Round-1 parallel count & 1{,}623 & 20.1 \\
\quad Round-1 query content & 439 & 5.4 \\
\quad Round-2 parallel count & 302 & 3.7 \\
\quad Round-2 tool type & 267 & 3.3 \\
\quad Round-2 query content & 101 & 1.2 \\
\quad Subtle wording & 823 & 10.2 \\
\addlinespace
\textit{Reasoning (all steps)} & \textit{1{,}374} & \textit{17.0} \\
\quad Pre-retrieval (step 2) & 886 & 10.9 \\
\quad Post-retrieval-1 (step 3) & 366 & 4.5 \\
\quad Post-retrieval-2 (step 4) & 122 & 1.5 \\
\addlinespace
Trace depth & 990 & 12.2 \\
Generation & 1{,}454 & 18.0 \\
Miscellaneous & 243 & 3.0 \\
\midrule
Total divergent pairs & 8{,}092 & 100.0 \\
\bottomrule
\end{tabular}
\caption{Full distribution of divergence types over all divergent pairs.}
\label{tab:divergence_full}
\end{table}


\begin{wraptable}{r}{0.48\columnwidth}
\centering
\small
\begin{tabular}{p{0.60\linewidth}rr}
\toprule
\textbf{Sub-category} & \textbf{Count} & \textbf{\%} \\
\midrule
Completeness difference & 840 & 57.8 \\
Ambiguous & 429 & 29.5 \\
Contradictory facts & 185 & 12.7 \\
\bottomrule
\end{tabular}
\caption{Sub-categories of generation-level divergent pairs.}
\label{tab:genbreak}
\end{wraptable}
\textbf{Generation-level divergence:} The generation category contains $1{,}454$ divergent pairs in which the retrieval
pipeline is identical but the final responses differ. We further classify these
pairs using GPT-4o~\citep{openai2024gpt4ocard} into completeness differences,
ambiguous cases, and contradictory factual claims. Of these pairs, $840$ differ
in completeness, $429$ are ambiguous, and $185$ contain contradictory factual
claims.


\section{Illustrative Examples}
\label{app:examples}

To preserve anonymity, product names in the following examples are replaced with
Product~A and Product~B, and the router-loaded configurations are shown by their
internal names in brackets.

\textbf{Queries with no configuration loaded:} For a small set of queries the
router loaded no domain configuration, and the agent answered from the model's own
parametric knowledge without retrieval. These are either generic technical questions
that can be answered without documentation, or queries for which the router failed to
identify a matching configuration. Representative examples are listed below.
\begin{itemize}\small
\item \textit{Does the schema support this date value ``2024-02-15Z20:20:39+00:00''?}
\item \textit{how do i test an access token using curl?}
\item \textit{what content type does the Product~A API require?}
\item \textit{whats an example of a sparkSession read?}
\item \textit{how do i create a local dataframe using pyspark 3?}
\item \textit{how do i create a local dataframe using scala?}
\item \textit{what is an ERD?}
\item \textit{My ingestion failed due to validation on an attribute, but I have that
attribute correctly in my file. What exactly is wrong?}
\end{itemize}

\textbf{Unstable routing:} Some queries loaded different configurations across
their ten runs.
\begin{itemize}\small
\item \textit{can we use policy service api to delete labels?}
  [\texttt{Product A}, \texttt{Product B}]
\item \textit{how to calculate min, max and sum for a journey?}
\item \textit{what are all the possible metric names I can query in observability
  metrics?} 
\item \textit{what do i do about the error ``Model Quality is poor''?}
\end{itemize}


\textbf{Divergence by pre-retrieval reasoning:}
\begin{quote}\small
\textbf{Query:} \emph{in Product~B, what are dispatcher fields}.
\textbf{Run A reasoning:} ``The user is asking about dispatcher fields. Let me search
the documentation.''
\textbf{Run B reasoning:} empty; Run B skipped the reasoning block.
\end{quote}

\paragraph{Divergence by trace depth.}
\begin{quote}\small
\textbf{Query:} \emph{is Product~A a data management platform?}
\textbf{Run A:} three steps (loaded configuration, searched, reasoned, answered).
\textbf{Run B:} one step (loaded configuration, answered directly without retrieval).
\end{quote}

\textbf{Divergence by generation:}
\begin{quote}\small
\textbf{Query:} \emph{what is a development sandbox?}
\textbf{Run A:} ``production sandboxes cannot be reset''.
\textbf{Run B:} ``production sandboxes can be reset, with caution''.
\end{quote}

\section{Search-Continuation Behavior}
\label{app:search}

The agent typically retrieves the ground-truth page within the first two retrieval
steps, yet continues searching after the hit in $89.8\%$ of runs. These additional
searches do not change the retrieved GT URL (Table~\ref{tab:search}).

\begin{table}[!htbp]
\centering\small
\begin{tabular}{lr}
\toprule
Mean first GT-hit step & 1.43 \\
GT found at step 1 / step 2 (\%) & 68.3 / 23.9 \\
Continued searching after the hit, per query (\%) & 96.7 \\
Continued searching after the hit, per run (\%) & 89.8 \\
\bottomrule
\end{tabular}
\caption{Frequency with which the agent continues searching after it has already
retrieved the ground-truth page.}
\label{tab:search}
\end{table}

\end{document}